\pdfoutput=1

\documentclass[11pt]{article}

\usepackage{ACL2023}

\usepackage{graphicx, subfigure}
\usepackage{tabularx}
\usepackage{CJKutf8}
\usepackage{color}
\usepackage{multirow}
\usepackage{stfloats}
\usepackage{array}

\usepackage{hyperref}

\usepackage{amsmath}

\usepackage{booktabs}

\usepackage[T1]{fontenc}

\usepackage{subfigure}

\usepackage{times}
\usepackage{latexsym}

\usepackage[T1]{fontenc}

\usepackage[utf8]{inputenc}

\usepackage{microtype}

\usepackage{inconsolata}

\title{Multi-Stage Coarse-to-Fine Contrastive Learning for \\ Conversation Intent Induction}

\author{
Caiyuan Chu$^1$\thanks{\hspace{1.5mm}Work done during the internship at iFLYTEK Research.} \thanks{\hspace{1.5mm}Equal contribution.}, 
Ya Li$^2$\footnotemark[2] \thanks{\hspace{1.5mm}Corresponding author.}, 
Yifan Liu$^2$, Jia-Chen Gu$^4$, Quan Liu$^{2,3}$, Yongxin Ge$^1$, Guoping Hu$^{2,3}$\\
  $^1$Chongqing University, Chongqing, China \\
  $^2$iFLYTEK Research, Hefei, China ~ 
  $^3$State Key Laboratory of Cognitive Intelligence ~ \\
  $^4$University of Science and Technology of China, Hefei, China \\
{\tt \{Chucy,yongxinge\}@cqu.edu.cn}, {\tt gujc@ustc.edu.cn}, \\
{\tt \{yali84,yfliu7,quanliu,gphu\}@iflytek.com}
}

\begin{document}
\maketitle

\begin{abstract}
Intent recognition is critical for task-oriented dialogue systems. However, for emerging domains and new services, it is difficult to accurately identify the key intent of a conversation due to time-consuming data annotation and comparatively poor model transferability. Therefore, the automatic induction of dialogue intention is very important for intelligent dialogue systems. This paper presents our solution to Track 2 of Intent Induction from Conversations for Task-Oriented Dialogue at the Eleventh Dialogue System Technology Challenge (DSTC11). The essence of intention clustering lies in distinguishing the representation of different dialogue utterances. The key to automatic intention induction is that, for any given set of new data, the sentence representation obtained by the model can be well distinguished from different labels. Therefore, we propose a multi-stage coarse-to-fine contrastive learning model training scheme including unsupervised contrastive learning pre-training, supervised contrastive learning pre-training, and fine-tuning with joint contrastive learning and clustering to obtain a better dialogue utterance representation model for the clustering task. In the released DSTC11 Track 2 evaluation results, our proposed system ranked first on both of the two subtasks of this Track.
\end{abstract}

\section{Introduction}
\begin{figure*}[t]
\centering
\includegraphics[width=1\textwidth]{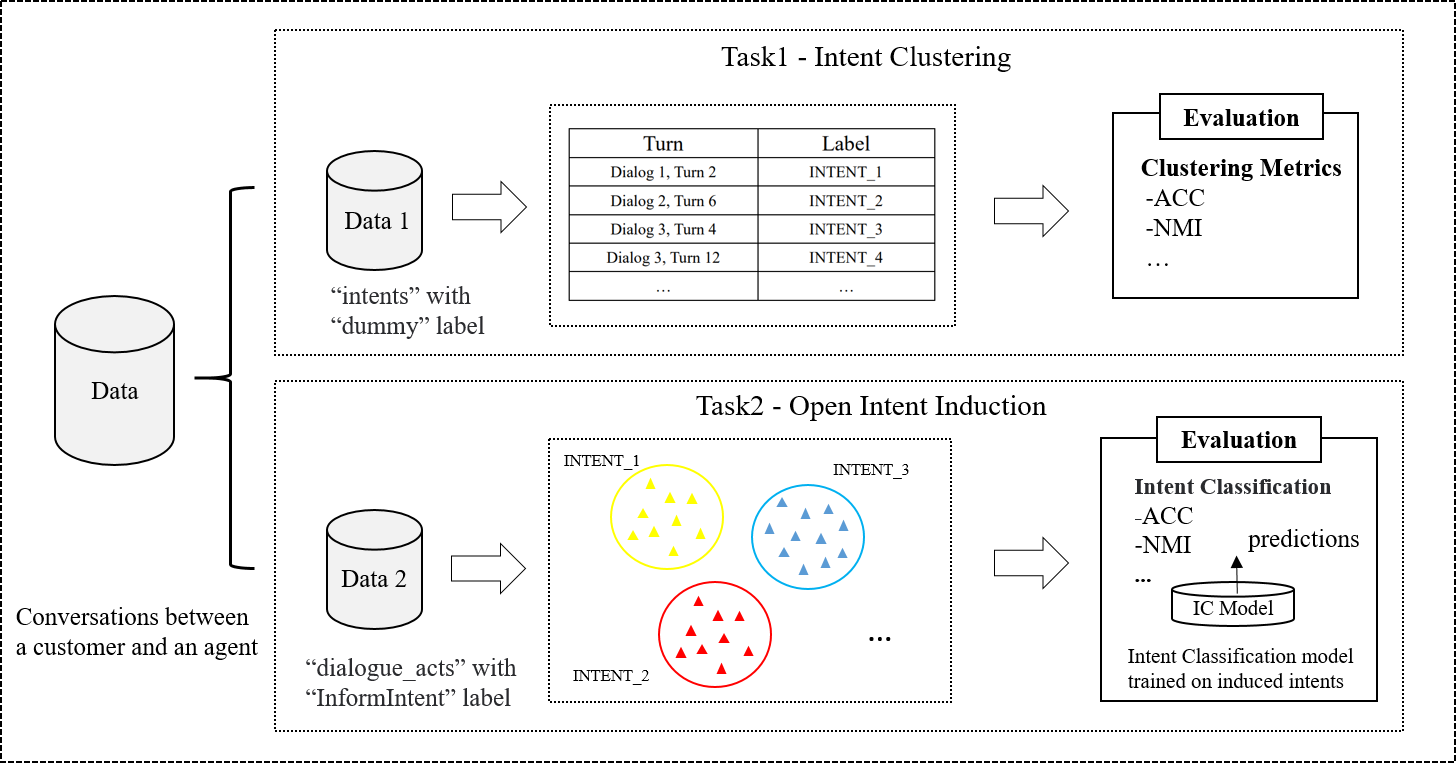} 
\caption{An overview of the tasks in this Track, including the tasks of Intent Clustering and Open Intent Induction.}
\label{fig1}
\end{figure*}

The design of dialogue mode is very important for the development of task-oriented dialogue system. It typically consists of a set of intents with corresponding slots for capturing and handling domain-specific dialogue box states. Previous work on schema-guided dialogue \citep{DBLP:journals/corr/abs-2002-01359,rastogi2020towards,DBLP:workshop/dstc/abs-2002-00181,lee2022sgd} focused on data-efficient joint dialogue state modeling across domains and zero-shot generalization to new APIs. However, for new emerging domains and novel services, the identification of key intents of such schema typically requires domain expertise and/or laborious analysis of a large volume of conversation transcripts. As the demand for and adoption of virtual assistants continues to increase, recent work has investigated ways to accelerate pattern design through the automatic induction of intentions~\citep{hakkani2015clustering,haponchyk2018supervised,DBLP:conf/emnlp/PerkinsY19, DBLP:journals/corr/abs-2005-11014} or the induction of slots and dialogue states~\citep{DBLP:conf/ijcai/MinQTL020,hudevcek2021discovering}. However, the lack of realistic shared benchmarks with public datasets, metrics, and task definitions has made it difficult to track progress in this area. 
For this reason, the Eleventh Dialogue System Technology Challenge (DSTC11) proposed a track of Intent Induction from Conversations for Task-Oriented Dialogue.
This Track is composed of two subtasks, which are are organized as shown in Fig.~\ref{fig1}.
1) Intent Clustering, which needs to cluster the given dialogue statements and evaluate them using standard clustering metrics. 2) Open Intent Induction, in which participants are required to generate a set of intents, each represented by a list of sample utterances. The induced intents and utterances will be evaluated using their performance on a downstream classification task over reference intents. Both tasks aim to discover intents from conversations. This task is very challenging due to the lack of labeled data and the unknown number of conversation intents.

This paper presents a system that is evaluated in this Track. For these tasks, in order to obtain a better dialogue utterance representation model under the condition that the data is unlabeled, we propose a multi-stage coarse-to-fine contrast learning model training scheme. The backbone of our multi-stage training scheme is the RoBERTa-large \citep{DBLP:journals/corr/abs-1907-11692}. Firstly, pre-training is performed by unsupervised contrastive learning in a large number of consecutive conversations. Secondly, the model is fine-tuned using a dataset with labels from the same domain as that of the target data for supervised comparative learning. Finally, the model obtained after the above two training steps is further fine-tuned on the target data by joint clustering and contrastive learning to obtain the final model. In addition, for the selection of the number of clustering categories k, we adopted the automatic parameter optimization method based on silhouette coefficients \citep{rousseeuw1987silhouettes} provided in the baseline code. As shown in the released evaluation results, our proposed model ranked first on both subtasks. Furthermore, experimental results are analyzed by ablation tests. Finally, we draw conclusions and give an overview of our future work.

\section{Related Work}
Labeled data for task-oriented dialogue systems is often scarce because of the high cost of data annotation. Consequently, learning generic dialogue representations that effectively capture dialogue semantics at various granularities \citep{hou2020few, DBLP:journals/corr/abs-2004-10793,DBLP:conf/cikm/GuLL19, DBLP:conf/naacl/YuHZDPL21} lays a good foundation for handling a variety of downstream tasks \citep{vinyals2016matching,snell2017prototypical}. Contrastive learning has recently demonstrated promising results in the processing of natural language. Among which SimCSE \citep{DBLP:conf/emnlp/GaoYC21} and TOD-BERT \citep{DBLP:conf/emnlp/WuHSX20} get a very good performance on general texts and dialogues, respectively. 
DSE \citep{DBLP:conf/naacl/0001Z0DMAX22} set new state-of-the-art results on general dialogues.

SimCSE \citep{DBLP:conf/emnlp/GaoYC21} uses Dropout \citep{srivastava2014dropout} to construct positive pairs by passing a sentence through the encoder twice to generate two different embeddings. Despite the fact that SimCSE performs better than ordinary data augmentation that directly manipulates discrete text, it has proven to be a poor performer in the field of dialogue, which is confirmed in the DSE \citep{DBLP:conf/naacl/0001Z0DMAX22}. Moreover, TOD-BERT takes an utterance and the concatenation of all the previous utterances in the dialogue as a positive pair. Although showing promise on the same tasks, TOD-BERT's semantic granularity and data statistics in many other dialogue tasks differ from those evaluated in their paper. DSE learns from dialogues by taking consecutive utterances of the same dialogue as positive pairs for contrastive learning, and state-of-the-art results are obtained in several tasks, including intention classification. Recently, MTP-CLNN \citep{DBLP:conf/acl/ZhangZZ0L22} set new state-of-the-art results in the New Intent Discovery field. However, the supervised pre-training part of it requires the data to have a small number of labels to get better results. According to our experimental verification, the generalization ability of the pre-trained model in the first stage is poor for data with no labels at all. In addition, recent research results on short text clustering show that the combined training method based on clustering and contrastive learning, SCCL \citep{DBLP:conf/naacl/ZhangNWLZMNAX21} achieves very good results. However, a better pre-trained model for the clustering task as encoded by SCCL can further improve the results of clustering. To this end, the goal of our work in the first two stages is to continue pre-training language models to derive better representations for downstream tasks.

\begin{figure*}[t]
\centering
\includegraphics[width=1\textwidth]{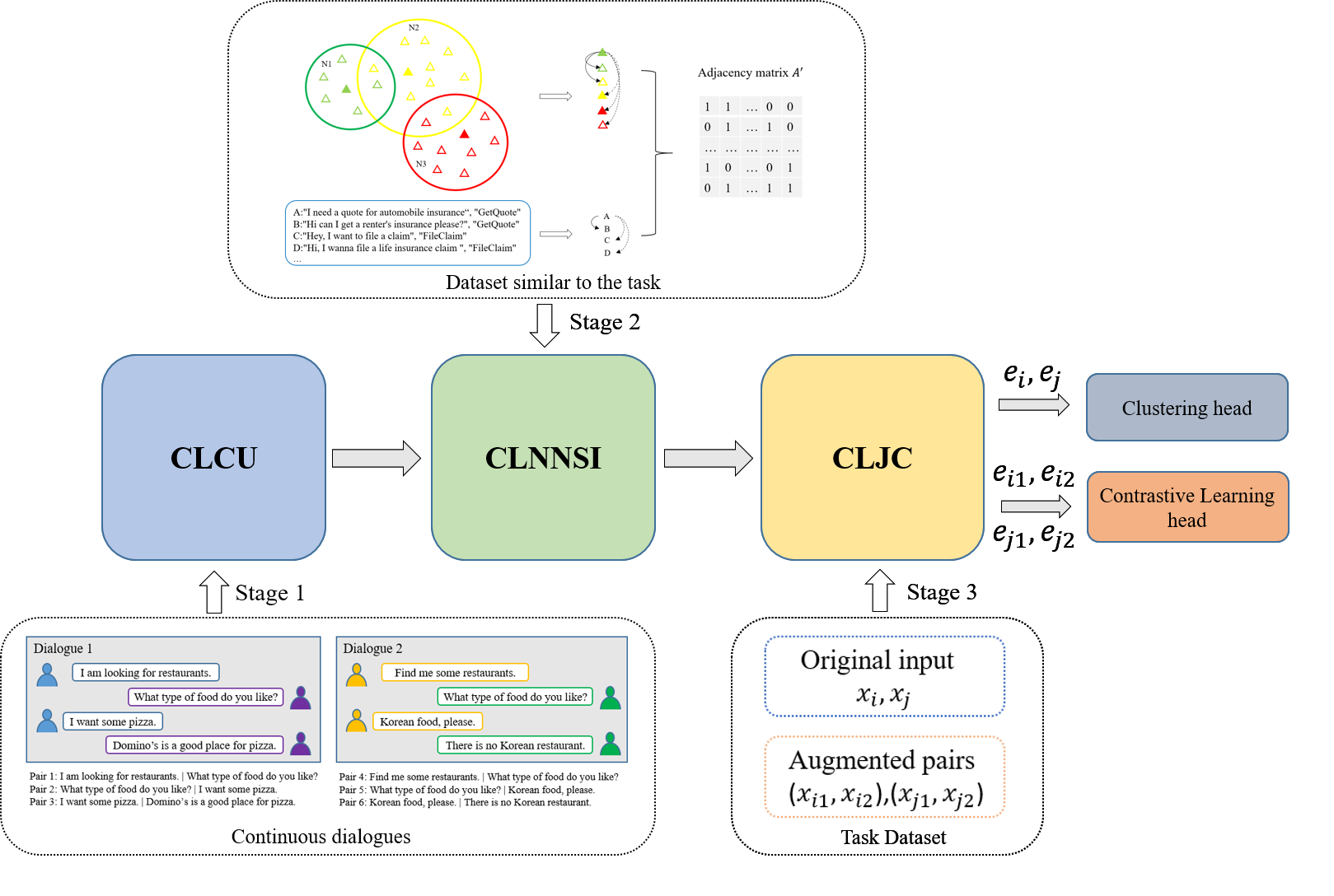} 
\caption{The framework of our proposed multi-stage coarse-to-fine contrastive learning model training scheme.}
\label{fig2}
\end{figure*}

\section{Methodology}
Our proposed multi-stage coarse-to-fine model training scheme consists of three stages: Contrastive Learning with Consecutive Utterances (CLCU), Contrastive Learning with the Nearest Neighbors and the Same Intent (CLNNSI) and Contrastive Learning with Joint Clustering (CLJC). This is shown in Fig.~\ref{fig2}. In the first stage, a pre-trained model is obtained by performing unsupervised contrastive learning on a large amount of dialogue data using consecutive discourses of the same dialogue as positive pairs. In the second stage, for labeled data in the same domain, we treat that sample with its neighboring samples or samples with the same intention as a positive pairs, and then fine-tune the model by contrastive learning. The third stage further fine-tunes the model by joint contrastive learning and clustering jointly on the target data. The negative pairs for contrastive learning are collected by small batches of negative sampling in the above three stages of model training. After training, we employ a simple non-parametric clustering algorithm named k-means to obtain clustering results. In this section, we describe our multi-stage coarse-to-fine contrastive learning model training scheme in detail below.

\subsection{Stage 1: CLCU}
CLCU encourages the model to treat an utterance as similar to its neighboring utterances and dissimilar to utterances that are not subsequent to it or that belong to other dialogues when doing contrastive learning on consecutive utterances. Consecutive utterances contain implicit categorical information, which benefits dialogue classification tasks (e.g., intent classification). Consider pairs 1 and 4 in Fig.~\ref{fig2} stage 1: We implicitly learn similar representations of \emph{I am looking for restaurants} and \emph{Find me some restaurants}, since they are both consecutive with \emph{What type of food do you like?}. In contrast, SimCSE \citep{DBLP:conf/emnlp/GaoYC21} does not enjoy these benefits by simply using Dropout as data augmentation. Although TOD-BERT also leverages the intrinsic semantics of dialogue by combining an utterance with its dialogue context as a positive pair, the context is often a concatenation of 5 to 15 utterances. Due to the large discrepancy in both semantics and data statistics between each utterance and its context, simply optimizing the similarity between them leads to less satisfying representations on many dialogue tasks. Just like the experimental results in DSE \citep{DBLP:conf/naacl/0001Z0DMAX22}, TOD-BERT can even lead to degenerated representations on some downstream tasks when compared to the original BERT~\citep{devlin-etal-2019-bert} model. Therefore, in the first stage, we adopt the method of DSE, which learns from dialogues by taking consecutive utterances of the same dialogue as positive pairs for contrastive learning, and directly use the model they trained on a large number of datasets as our first stage model.

\subsection{Stage 2: CLNNSI} 

In the second stage, a small number of labeled datasets in the same domain as the target data are used, which makes our model have stronger generalization ability. And adopt an objective that pulls neighboring instances together and pushes distant ones away in the embedding space to learn compact representations for clustering. To be specific, we first encode the utterances with the pre-trained model from stage 1. The inner product is then used as a distance metric to find the top-K nearest neighbors of each utterance $x_i$ in the embedding space, forming a neighborhood $N_i$. During training, for each minibatch of utterances $\mathrm{B}=\left\{x_{i}\right\}_{i=1}^{M}$ and each utterance $x_{i} \in \mathrm{B}$, we uniformly sample one neighbor $x_{i}^{\prime}$ from its neighborhood $N_i$. Then use data augmentation to generate $\tilde{x}_{i}$ and $\tilde{x}_{i}^{\prime}$ for $x_i$ and $x_{i}^{\prime}$ respectively. Here, $\tilde{x}_{i}$ and $\tilde{x}_{i}^{\prime}$ are treated as a positive pair. We then obtain an augmented batch $B^{\prime}=\left\{\tilde{x}_{i},\tilde{x}_{i}^{\prime}\right\}_{i=1}^{M}$  with all the generated samples. To compute contrastive loss, we construct an adjacency matrix $A^{\prime}$  for $B^{\prime}$, which is a 2M × 2M binary matrix where 1 indicates positive relations (either being neighbors or having the same intent label) and 0 indicates negative relations. Hence, the contrastive loss can writed as: 
\begin{equation}
l_{i}=-\frac{1}{\left|\mathcal{C}_{i}\right|} \sum_{j \in C_{i}}{ \log{\frac{\exp{ \left(\operatorname{sim}\left(\tilde{e}_{i}, \tilde{e}_{j}\right) / \tau\right)}}{\sum_{k \neq i}^{2M}{ \exp{\left(\operatorname{sim}\left(\tilde{e}_{i}, \tilde{e}_{k}\right) / \tau\right)}}} } },
\end{equation}
\begin{equation}
L_{s t g 2}=\frac{1}{2 M} \sum_{i=1}^{2 M} l_{i},
\end{equation}
where $C_{i}=\left\{A_{i j}^{\prime}=1 \mid j \in\{1, \ldots, 2 M\}\right\}$ denotes the set of instances having positive relation with $\tilde{x}_{i}$ and $\left|C_{i}\right|$ is the cardinality. $\tilde{e}_{i}$ is the embedding for utterance $\tilde{x}_{i}$, $\tau$ is the temperature parameter. $sim(.,.)$ is a similarity function on a pair of normalized feature vectors. Has the following advantages by introducing the notion of neighborhood relationships in contrastive learning: 1) Similar instances are pulled together and dissimilar instances are pushed away to achieve more compact clusters; and 2) known intents are naturally incorporated with the adjacency matrix.

\begin{figure}[t]
\centering
\includegraphics[scale=0.45]{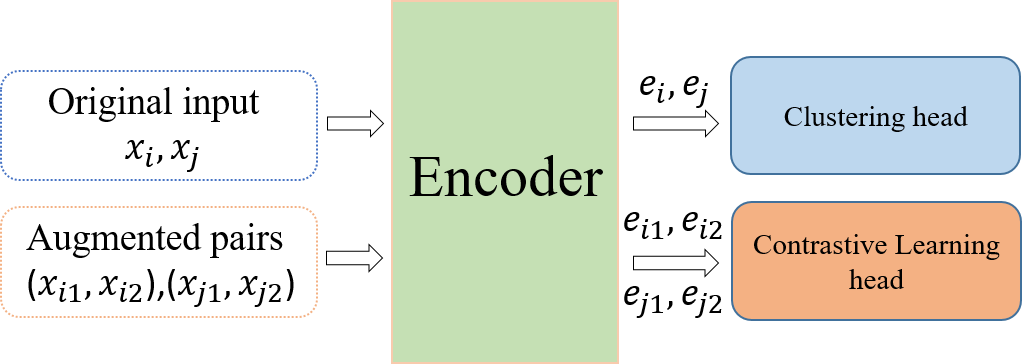}  
\caption{The training framework for stage 3.}  
\label{fig3}    
\end{figure}

\subsection{Stage 3: CLJC}
Previous research efforts focused on integrating clustering with deep representation learning by optimizing a clustering objective defined in the representation space \citep{zhang2017deep,DBLP:conf/iclr/ShahamSLBNK18}. Despite promising improvements, the clustering performance is still inadequate, especially in the presence of complex data with a large number of clusters. One possible reason is that, even with a deep neural network, data still has significant overlap across categories before clustering starts. Consequently, the clusters learned by optimizing various distance or similarity-based clustering objectives suffer from poor purity. Moreover, contrastive learning has recently achieved remarkable success in self-monitoring \citep{wu2018unsupervised,chen2020simple}, as the name suggests, a contrastive loss is adopted to pull together samples augmented from the same instance in the original dataset while pushing apart those from different ones. This beneficial property can be leveraged to support clustering by scattering apart the overlapped categories. Hence, in the third stage, for target data, we adopt the method of joint clustering and contrastive learning to further fine-tune the model. The training framework stage3 is shown in Fig.~\ref{fig3}. Among them, $x_{i_{1}}$,$x_{i_{2}}$ and $x_{j_{1}}$,$x_{j_{2}}$ are obtained by means of data augmentation. In this paper, we adopt the contextual augmenter data augment in the form of a pre-trained transformers to find the top-n suitable words of the input text for insertion or substitution. We used word substitution to augment the data and chose Bertbase and Robertabase to generate augmented pairs. And the overall objective is:
\begin{equation}
L_{s t g 3}={L}_{C L}+\eta{L}_{C l u},
\end{equation}
where ${L}_{C L}$ and ${L}_{C l u}$ are the loss functions of comparative learning and clustering, respectively. $\eta$ balances between the contrastive loss and the clustering loss of stage3, for simplicity, it is set to 10 in our experiment.

\section{Experiments}
\subsection{Dataset}
For this Track, one development dataset and two test datasets are provided. Each dataset consists of 1) some human-to-human conversations between a customer and an agent, and 2) a set of testing samples with corresponding intent annotations. The task1-utterances and task2-utterances are utterances in which "intents" are non-empty and "dialogue\_acts" are "InformIntent" in dialogues in each dataset, respectively. Detailed statistics of the dataset are summarized in Table~\ref{tab:1}.

\begin{table}[t]
	\centering
	\begin{tabular}{ccccc}
		\hline\hline\noalign{\smallskip}	
        dataset & dial. & test-utt. & task1-utt. & task2-utt.  \\
		\noalign{\smallskip}\hline\noalign{\smallskip}
		dev & 948 & 913 & 1205 & 4332   \\
		banking & 1000 & 407 & 1503 & 3696  \\
		finance & 2000 & 1130 & 1597 & 6676  \\
		\noalign{\smallskip}\hline
	\end{tabular}
        
    \caption{Statistics of the DSTC11-Track 2 datasets. utt.: utterance, dial.: dialogue.}
    \label{tab:1}  
\end{table}

\subsection{Metrics}
Task 1 and task 2 are both evaluated by the following six metrics,: accuracy (ACC) \citep{huang2014deep}, normalized mutual information (NMI), F1-score, Recall, Precision, and adjusted rand index (ARI). But the ACC is the primary metric used for ranking system submissions. Metrics dependent on reference intents will be computed using an automatic alignment of cluster labels to reference intent labels. For task 1, alignments will be computed based on turn-level reference intent labels. For task 2, to avoid the need to assign labels to turns in the input transcripts, alignments will be computed using classifier predictions on the set of utterances held out for evaluation. In both cases, 1:1 alignments between induced intents and reference intents will be computed using the Hungarian algorithm \citep{kuhn1955hungarian}.

\begin{table}[t]
	\centering
	\setlength{\tabcolsep}{2.8pt}
	\begin{tabular}{lcccccc}
		\hline\hline\noalign{\smallskip}	
		Team& ACC& P& R& F1& NMI& ARI \\
		\noalign{\smallskip}\hline\noalign{\smallskip}
		\textbf{T23}&\textbf{69.79}&	76.09&	76.12&	76.00&	75.05	&59.23 \\
        T07& 69.59&	72.01&	81.64&	\textbf{76.50}&	73.48&	60.13\\
        T35& 69.31&	\textbf{78.41}&	73.35&	75.78&	74.30&	58.67\\
        T05& 69.06&	73.26&	78.32&	75.70&	\textbf{75.54}&	\textbf{61.38}\\
        T02& 68.83&	73.04&	78.06&	75.46&	75.13&	60.88\\
        T17& 67.15&	70.49&	78.48&	74.10&	73.20&	60.86\\
        T36& 66.32&	71.86&	74.46&	73.13&	73.64&	58.05\\
        T24& 66.19&	71.17&	77.36&	74.13&	74.72&	58.15\\
        T00& 64.92&	71.42&	74.80&	73.05&	71.08&	50.37\\
        T34&63.73&	71.01&	74.84&	72.59&	72.98&	52.77\\
        …&	…&	…&	…&	…&	…& …  \\
        baseline& 55.80&	64.97&	62.98&	63.26&	62.98&	39.85\\
		\noalign{\smallskip}\hline
	\end{tabular}
    \caption{The summary of the task 1 evaluation results on the two datasets, with the best score in bold.}
    \label{tab:2}
\end{table}

\begin{table}[t]
	\centering
	\setlength{\tabcolsep}{2.8pt}
	\begin{tabular}{lcccccc}
		\hline\hline\noalign{\smallskip}	
		Team& ACC& P& R& F1& NMI& ARI\\
		\noalign{\smallskip}\hline\noalign{\smallskip}
        \textbf{T23} &\textbf{76.30}&78.68&\textbf{89.86}&83.55&87.42&\textbf{71.82}\\
        T02&75.34&78.18&88.18&82.86&87.32&68.87 \\
        T36&74.85&78.81&87.59&82.85&87.00&71.36\\
        T24&74.70&79.76&87.62&83.42&87.45&70.71\\
        T05&74.52&79.50&87.49&83.17&87.88&70.26\\
        T17&73.79&\textbf{83.25}&84.87&\textbf{83.99}&\textbf{88.11}&71.42\\
        T14&69.55&70.68&87.97&78.27&83.67&65.11\\
        T13&69.43&82.66&74.47&78.26&82.29&62.13\\
        T27&68.70&80.29&77.51&78.76&83.69&63.83\\
        T19& 67.50&69.09&89.4&77.92&83.73&63.67\\
        …&…&…&…&…&…&…\\
        baseline&63.61&68.93&79.59&73.86&80.09&57.24\\
		\noalign{\smallskip}\hline
	\end{tabular}
	\caption{The summary of the task 2 evaluation results on the two datasets, with the best score in bold.}
	\label{tab:3} 
\end{table}

\subsection{Experiment Results}
The final summary results of the test-banking dataset and the test-finance dataset for the two tasks are shown in Tables~\ref{tab:2} and~\ref{tab:3}. Our model ranks first in the test sets for both subtasks. Compared with the baseline, where the baseline model in the above tables is the official provided baseline model mpnet by Track 2 of DSTC11. It is trained on a large and diverse dataset of over 1 billion training pairs. And the test ACC of our model is improved by 13.99\% and 12.69\% on tasks 1 and 2, respectively. It shows the effectiveness of our method. 

In addition, to test the advantages of the proposed method on other dataset in this paper, we give the results of our method on the development dataset compared with some representative baseline models. We compared with the following baseline models: \textbf{(1) GloVe}~\citep{DBLP:conf/emnlp/PenningtonSM14}. One of the official baseline models provided by the Track 2 of DSTC11, which is a sentence-transformers model. It maps sentences or paragraphs to a 300 dimensional dense vector space and can be used for tasks like clustering or semantic search. 
\textbf{(2) MPNet}~\citep{DBLP:conf/nips/Song0QLL20}. Another official baseline model provided by Track 2 of DSTC11, which is trained on a large and diverse dataset of over 1 billion training pairs. The base model is MPNet-base and the dimension is 768.
\textbf{(3) SimCSE} \citep{DBLP:conf/emnlp/GaoYC21}. The model we chose is trained on 10$^6$ randomly sampled sentences from English Wikipedia by unsupervised contrastive learning, and its base model is RoBERTa-large with a dimension of 1024.
\textbf{(4) DSE} \citep{DBLP:conf/naacl/0001Z0DMAX22}. It is trained on a large number of training pairs constructed in the style of consecutive conversational sentences as positive pair using multiple conversational datasets. The base model is RoBERTa-large with a dimension of 1024.
\textbf{(5) SCCL} \citep{DBLP:conf/naacl/ZhangNWLZMNAX21}. Fine-tuning of the baseline model using the development  dataset by joint training with clustering and contrastive learning, where the baseline model is the DSE-trained model described above.

\begin{table}[t]
	\centering
	\begin{tabular}{lcc}
		\hline\hline\noalign{\smallskip}	
          & task1	&task2  \\
		\noalign{\smallskip}\hline\noalign{\smallskip}

        GloVe\citep{DBLP:conf/emnlp/PenningtonSM14} &	20.58&	29.12 \\
        MPNet \citep{DBLP:conf/nips/Song0QLL20} 
        &	46.14&	60.23 \\
        SimCSE \citep{DBLP:conf/emnlp/GaoYC21} 
        &   47.39&	57.74\\
        DSE \citep{DBLP:conf/naacl/0001Z0DMAX22} 
        &	58.67&	64.18 \\
        SCCL \citep{DBLP:conf/naacl/ZhangNWLZMNAX21} &	65.32&	78.34 \\
        \textbf{Our model}&	\textbf{75.68}&	\textbf{85.76} \\
  
		\noalign{\smallskip}\hline
	\end{tabular}
	\caption{Comparison of the accuracy of different models on the development dataset for the two tasks. }
	\label{tab:4} 
\end{table}

As can be seen from the experimental results in Table~\ref{tab:4}, our model obtained the best results in both subtasks compared to several other representative baseline models, with accuracy rates of 75.68\% and 85.76\%, respectively. On task 1, the accuracy of our model outperformed GloVe by 55.1\%, outperformed MPNet by 29.54\%, outperformed SimCSE by 28.29\%, outperformed DSE by 17.01\%, and outperformed SCCL by 10.36\%. On task 2, the accuracy of our model outperformed GloVe by 56.64\%, outperformed MPNet by 25.53\%, outperformed SimCSE by 28.02\%, outperformed DSE by 21.56\%, and outperformed SCCL by 7.42\%. This illustrates the soundness of our approach.

\subsection{Ablation Study}
\begin{figure*}[t]
\centering
\subfigure[Baseline]{\label{fig:subfig:a}
\includegraphics[scale=0.45]{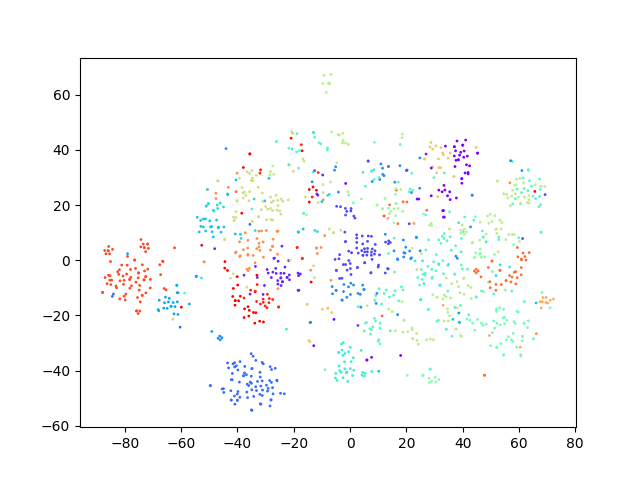}}
\hspace{0.005\linewidth}
\subfigure[Stage 1]{\label{fig:subfig:b}
\includegraphics[scale=0.45]{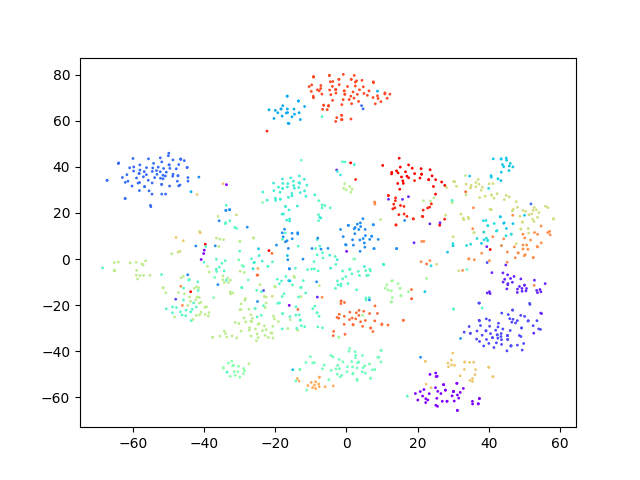}}
\vfill
\subfigure[Stage 2]{\label{fig:subfig:c}
\includegraphics[scale=0.45]{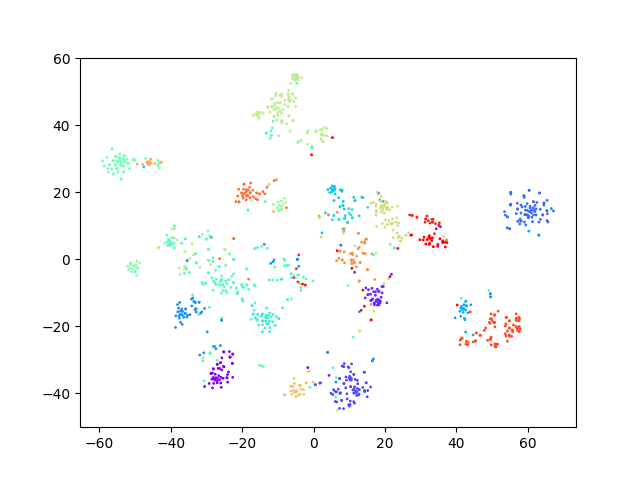}}
\hspace{0.005\linewidth}
\subfigure[Stage 3]{\label{fig:subfig:d}
\includegraphics[scale=0.45]{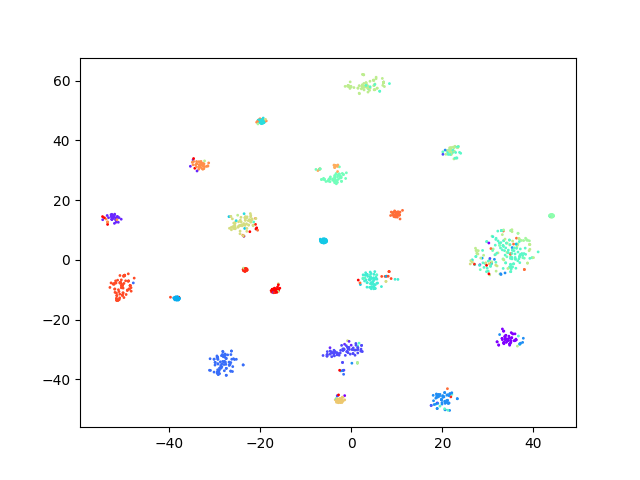}}
\label{fig:subfig}
\caption{
TSNE visualization \citep{van2008visualizing} of the dialogue representations provides by baseline and three stages on the development set, each color indicates a ground truth semantic category.
}
\label{fig4}
\end{figure*}

Through ablation experiments with a random seed set to 42 in the clustering algorithm on the development dataset, the experimental results are shown in Table~\ref{tab:5}. When there is no stage1, stage2 and stag3 (this is the case is the baseline), the ACC of the test on task1 and task2 are 46.14\%, 60.24\% respectively. when stage1, stage2 and stage3 are performed, the results on task1 are improved by 12.45\%, 12.12\% and the results on task2 improved by 0.68\%, 19.38\%, and 5.48\%, respectively. In addition, the TSNE visualization of the utterances representation of the model obtained at each stage on the development set is shown in Fig.~\ref{fig4}. It can be observed from the figure that each cluster becomes more and more compact after each stage compared to the baseline model. That verifies the rationality of each stage.

\begin{table}[t]
	\centering
	\begin{tabular}{lcc}
		\hline\hline\noalign{\smallskip}	
          & task1	&task2  \\
		\noalign{\smallskip}\hline\noalign{\smallskip}
		model&	75.68&	85.76 \\
		~~~w/o. stage1,stage2 and stage3 &	46.14&	60.24   \\
		~~~w/o. stage2 and stage3&	58.59&	60.90  \\
		~~~w/o. stage3&	70.71&	80.28  \\
		
		\noalign{\smallskip}\hline
	\end{tabular}
	\caption{Accuracy of ablation experiment on the development set.}
	\label{tab:5} 
        \vspace{-3mm}
\end{table}

\section{Conclusion}
In this paper, we present our solution to the challenge of Intent Induction from Conversations for task-Oriented Dialogue of DSTC11. Firstly, we chose RoBERTa-large, which was pre-trained on a large number of continuous dialogues. Since continuous utterances also contain implicit classification information, they are beneficial for the task of dialogue intention classification. Secondly, the use of some labeled datasets in the same domain as the target data makes our model have stronger transfer ability. And we adopt KNN to select the positive pair for contrastive learning in order to pull neighboring instances together and push distant ones away in the embedding space to learn compact representations for clustering. Lastly, the joint training method of clustering and contrastive learning makes the advantages of clustering and contrastive learning complementary. Experimental results demonstrate that our methods can effectively cluster the utterances with intention in the dialogue. Our method of competitive performance achieved first place in two subtasks. In the future, we will explore better ways to obtain a better dialogue utterance representation model for the clustering task.

\section*{Acknowledgements}
We thank anonymous reviewers for their valuable comments.

\bibliography{acl2023}

\begin{thebibliography}{34}
\expandafter\ifx\csname natexlab\endcsname\relax\def\natexlab#1{#1}\fi

\bibitem[{Chatterjee and Sengupta(2020)}]{DBLP:journals/corr/abs-2005-11014}
Ajay Chatterjee and Shubhashis Sengupta. 2020.
\newblock \href {http://arxiv.org/abs/2005.11014} {Intent mining from past
  conversations for conversational agent}.
\newblock \emph{CoRR}, abs/2005.11014.

\bibitem[{Chen et~al.(2020)Chen, Kornblith, Norouzi, and
  Hinton}]{chen2020simple}
Ting Chen, Simon Kornblith, Mohammad Norouzi, and Geoffrey Hinton. 2020.
\newblock A simple framework for contrastive learning of visual
  representations.
\newblock In \emph{International conference on machine learning}, pages
  1597--1607. PMLR.

\bibitem[{Devlin et~al.(2019)Devlin, Chang, Lee, and
  Toutanova}]{devlin-etal-2019-bert}
Jacob Devlin, Ming-Wei Chang, Kenton Lee, and Kristina Toutanova. 2019.
\newblock \href {https://doi.org/10.18653/v1/N19-1423} {{BERT}: Pre-training of
  deep bidirectional transformers for language understanding}.
\newblock In \emph{Proceedings of the 2019 Conference of the North {A}merican
  Chapter of the Association for Computational Linguistics: Human Language
  Technologies, Volume 1 (Long and Short Papers)}, pages 4171--4186,
  Minneapolis, Minnesota. Association for Computational Linguistics.

\bibitem[{Gao et~al.(2021)Gao, Yao, and Chen}]{DBLP:conf/emnlp/GaoYC21}
Tianyu Gao, Xingcheng Yao, and Danqi Chen. 2021.
\newblock \href {https://doi.org/10.18653/v1/2021.emnlp-main.552} {Simcse:
  Simple contrastive learning of sentence embeddings}.
\newblock In \emph{Proceedings of the 2021 Conference on Empirical Methods in
  Natural Language Processing, {EMNLP} 2021, Virtual Event / Punta Cana,
  Dominican Republic, 7-11 November, 2021}, pages 6894--6910. Association for
  Computational Linguistics.

\bibitem[{Gu et~al.(2019)Gu, Ling, and Liu}]{DBLP:conf/cikm/GuLL19}
Jia{-}Chen Gu, Zhen{-}Hua Ling, and Quan Liu. 2019.
\newblock Interactive matching network for multi-turn response selection in
  retrieval-based chatbots.
\newblock In \emph{Proceedings of the 28th {ACM} International Conference on
  Information and Knowledge Management, {CIKM} 2019}, pages 2321--2324. {ACM}.

\bibitem[{Hakkani-T{\"u}r et~al.(2015)Hakkani-T{\"u}r, Ju, Zweig, and
  Tur}]{hakkani2015clustering}
Dilek Hakkani-T{\"u}r, Yun-Cheng Ju, Geoffrey Zweig, and Gokhan Tur. 2015.
\newblock Clustering novel intents in a conversational interaction system with
  semantic parsing.
\newblock In \emph{Sixteenth Annual Conference of the International Speech
  Communication Association}.

\bibitem[{Haponchyk et~al.(2018)Haponchyk, Uva, Yu, Uryupina, and
  Moschitti}]{haponchyk2018supervised}
Iryna Haponchyk, Antonio Uva, Seunghak Yu, Olga Uryupina, and Alessandro
  Moschitti. 2018.
\newblock Supervised clustering of questions into intents for dialog system
  applications.
\newblock In \emph{Proceedings of the 2018 Conference on Empirical Methods in
  Natural Language Processing}, pages 2310--2321.

\bibitem[{Hou et~al.(2020)Hou, Che, Lai, Zhou, Liu, Liu, and Liu}]{hou2020few}
Yutai Hou, Wanxiang Che, Yongkui Lai, Zhihan Zhou, Yijia Liu, Han Liu, and Ting
  Liu. 2020.
\newblock Few-shot slot tagging with collapsed dependency transfer and
  label-enhanced task-adaptive projection network.
\newblock In \emph{Proceedings of the 58th Annual Meeting of the Association
  for Computational Linguistics}, pages 1381--1393.

\bibitem[{Huang et~al.(2014)Huang, Huang, Wang, and Wang}]{huang2014deep}
Peihao Huang, Yan Huang, Wei Wang, and Liang Wang. 2014.
\newblock Deep embedding network for clustering.
\newblock In \emph{2014 22nd International conference on pattern recognition},
  pages 1532--1537. IEEE.

\bibitem[{Hude{\v{c}}ek et~al.(2021)Hude{\v{c}}ek, Du{\v{s}}ek, and
  Yu}]{hudevcek2021discovering}
Vojt{\v{e}}ch Hude{\v{c}}ek, Ond{\v{r}}ej Du{\v{s}}ek, and Zhou Yu. 2021.
\newblock Discovering dialogue slots with weak supervision.
\newblock In \emph{Proceedings of the 59th Annual Meeting of the Association
  for Computational Linguistics and the 11th International Joint Conference on
  Natural Language Processing (Volume 1: Long Papers)}, pages 2430--2442.

\bibitem[{Krone et~al.(2020)Krone, Zhang, and
  Diab}]{DBLP:journals/corr/abs-2004-10793}
Jason Krone, Yi~Zhang, and Mona~T. Diab. 2020.
\newblock \href {http://arxiv.org/abs/2004.10793} {Learning to classify intents
  and slot labels given a handful of examples}.
\newblock \emph{CoRR}, abs/2004.10793.

\bibitem[{Kuhn(1955)}]{kuhn1955hungarian}
Harold~W Kuhn. 1955.
\newblock The hungarian method for the assignment problem.
\newblock \emph{Naval research logistics quarterly}, 2(1-2):83--97.

\bibitem[{Lee et~al.(2022)Lee, Gupta, Rastogi, Cao, Zhang, and Wu}]{lee2022sgd}
Harrison Lee, Raghav Gupta, Abhinav Rastogi, Yuan Cao, Bin Zhang, and Yonghui
  Wu. 2022.
\newblock Sgd-x: A benchmark for robust generalization in schema-guided
  dialogue systems.
\newblock In \emph{Proceedings of the AAAI Conference on Artificial
  Intelligence}, volume~36, pages 10938--10946.

\bibitem[{Liu et~al.(2019)Liu, Ott, Goyal, Du, Joshi, Chen, Levy, Lewis,
  Zettlemoyer, and Stoyanov}]{DBLP:journals/corr/abs-1907-11692}
Yinhan Liu, Myle Ott, Naman Goyal, Jingfei Du, Mandar Joshi, Danqi Chen, Omer
  Levy, Mike Lewis, Luke Zettlemoyer, and Veselin Stoyanov. 2019.
\newblock \href {http://arxiv.org/abs/1907.11692} {Roberta: {A} robustly
  optimized {BERT} pretraining approach}.
\newblock \emph{CoRR}, abs/1907.11692.

\bibitem[{Min et~al.(2020)Min, Qin, Teng, Liu, and
  Zhang}]{DBLP:conf/ijcai/MinQTL020}
Qingkai Min, Libo Qin, Zhiyang Teng, Xiao Liu, and Yue Zhang. 2020.
\newblock \href {https://doi.org/10.24963/ijcai.2020/532} {Dialogue state
  induction using neural latent variable models}.
\newblock In \emph{Proceedings of the Twenty-Ninth International Joint
  Conference on Artificial Intelligence, {IJCAI} 2020}, pages 3845--3852.
  ijcai.org.

\bibitem[{Pennington et~al.(2014)Pennington, Socher, and
  Manning}]{DBLP:conf/emnlp/PenningtonSM14}
Jeffrey Pennington, Richard Socher, and Christopher~D. Manning. 2014.
\newblock \href {https://doi.org/10.3115/v1/d14-1162} {Glove: Global vectors
  for word representation}.
\newblock In \emph{Proceedings of the 2014 Conference on Empirical Methods in
  Natural Language Processing, {EMNLP} 2014, October 25-29, 2014, Doha, Qatar,
  {A} meeting of SIGDAT, a Special Interest Group of the {ACL}}, pages
  1532--1543. {ACL}.

\bibitem[{Perkins and Yang(2019)}]{DBLP:conf/emnlp/PerkinsY19}
Hugh Perkins and Yi~Yang. 2019.
\newblock \href {https://doi.org/10.18653/v1/D19-1413} {Dialog intent induction
  with deep multi-view clustering}.
\newblock In \emph{Proceedings of the 2019 Conference on Empirical Methods in
  Natural Language Processing and the 9th International Joint Conference on
  Natural Language Processing, {EMNLP-IJCNLP} 2019, Hong Kong, China, November
  3-7, 2019}, pages 4014--4023. Association for Computational Linguistics.

\bibitem[{Rastogi et~al.(2020{\natexlab{a}})Rastogi, Zang, Sunkara, Gupta, and
  Khaitan}]{DBLP:journals/corr/abs-2002-01359}
Abhinav Rastogi, Xiaoxue Zang, Srinivas Sunkara, Raghav Gupta, and Pranav
  Khaitan. 2020{\natexlab{a}}.
\newblock \href {http://arxiv.org/abs/2002.01359} {Schema-guided dialogue state
  tracking task at {DSTC8}}.
\newblock \emph{CoRR}, abs/2002.01359.

\bibitem[{Rastogi et~al.(2020{\natexlab{b}})Rastogi, Zang, Sunkara, Gupta, and
  Khaitan}]{rastogi2020towards}
Abhinav Rastogi, Xiaoxue Zang, Srinivas Sunkara, Raghav Gupta, and Pranav
  Khaitan. 2020{\natexlab{b}}.
\newblock Towards scalable multi-domain conversational agents: The
  schema-guided dialogue dataset.
\newblock In \emph{Proceedings of the AAAI Conference on Artificial
  Intelligence}, volume~34, pages 8689--8696.

\bibitem[{Rousseeuw(1987)}]{rousseeuw1987silhouettes}
Peter~J Rousseeuw. 1987.
\newblock Silhouettes: a graphical aid to the interpretation and validation of
  cluster analysis.
\newblock \emph{Journal of computational and applied mathematics}, 20:53--65.

\bibitem[{Ruan et~al.(2020)Ruan, Ling, Gu, and
  Liu}]{DBLP:workshop/dstc/abs-2002-00181}
Yu{-}Ping Ruan, Zhen{-}Hua Ling, Jia{-}Chen Gu, and Quan Liu. 2020.
\newblock Fine-tuning {BERT} for schema-guided zero-shot dialogue state
  tracking.
\newblock In \emph{Proceedings of the Thirty-Fourth {AAAI} Conference on
  Artificial Intelligence, {AAAI} 2020, Workshop on the Eighth Dialog System
  Technology Challenge, {DSTC8}}.

\bibitem[{Shaham et~al.(2018)Shaham, Stanton, Li, Basri, Nadler, and
  Kluger}]{DBLP:conf/iclr/ShahamSLBNK18}
Uri Shaham, Kelly~P. Stanton, Henry Li, Ronen Basri, Boaz Nadler, and Yuval
  Kluger. 2018.
\newblock \href {https://openreview.net/forum?id=HJ\_aoCyRZ} {Spectralnet:
  Spectral clustering using deep neural networks}.
\newblock In \emph{6th International Conference on Learning Representations,
  {ICLR} 2018, Vancouver, BC, Canada, April 30 - May 3, 2018, Conference Track
  Proceedings}. OpenReview.net.

\bibitem[{Snell et~al.(2017)Snell, Swersky, and Zemel}]{snell2017prototypical}
Jake Snell, Kevin Swersky, and Richard Zemel. 2017.
\newblock Prototypical networks for few-shot learning.
\newblock \emph{Advances in neural information processing systems}, 30.

\bibitem[{Song et~al.(2020)Song, Tan, Qin, Lu, and
  Liu}]{DBLP:conf/nips/Song0QLL20}
Kaitao Song, Xu~Tan, Tao Qin, Jianfeng Lu, and Tie{-}Yan Liu. 2020.
\newblock \href
  {https://proceedings.neurips.cc/paper/2020/hash/c3a690be93aa602ee2dc0ccab5b7b67e-Abstract.html}
  {Mpnet: Masked and permuted pre-training for language understanding}.
\newblock In \emph{Advances in Neural Information Processing Systems 33: Annual
  Conference on Neural Information Processing Systems 2020, NeurIPS 2020,
  December 6-12, 2020, virtual}.

\bibitem[{Srivastava et~al.(2014)Srivastava, Hinton, Krizhevsky, Sutskever, and
  Salakhutdinov}]{srivastava2014dropout}
Nitish Srivastava, Geoffrey Hinton, Alex Krizhevsky, Ilya Sutskever, and Ruslan
  Salakhutdinov. 2014.
\newblock Dropout: a simple way to prevent neural networks from overfitting.
\newblock \emph{The journal of machine learning research}, 15(1):1929--1958.

\bibitem[{Van~der Maaten and Hinton(2008)}]{van2008visualizing}
Laurens Van~der Maaten and Geoffrey Hinton. 2008.
\newblock Visualizing data using t-sne.
\newblock \emph{Journal of machine learning research}, 9(11).

\bibitem[{Vinyals et~al.(2016)Vinyals, Blundell, Lillicrap, Wierstra
  et~al.}]{vinyals2016matching}
Oriol Vinyals, Charles Blundell, Timothy Lillicrap, Daan Wierstra, et~al. 2016.
\newblock Matching networks for one shot learning.
\newblock \emph{Advances in neural information processing systems}, 29.

\bibitem[{Wu et~al.(2020)Wu, Hoi, Socher, and Xiong}]{DBLP:conf/emnlp/WuHSX20}
Chien{-}Sheng Wu, Steven C.~H. Hoi, Richard Socher, and Caiming Xiong. 2020.
\newblock \href {https://doi.org/10.18653/v1/2020.emnlp-main.66} {{TOD-BERT:}
  pre-trained natural language understanding for task-oriented dialogue}.
\newblock In \emph{Proceedings of the 2020 Conference on Empirical Methods in
  Natural Language Processing, {EMNLP} 2020, Online, November 16-20, 2020},
  pages 917--929. Association for Computational Linguistics.

\bibitem[{Wu et~al.(2018)Wu, Xiong, Yu, and Lin}]{wu2018unsupervised}
Zhirong Wu, Yuanjun Xiong, Stella~X Yu, and Dahua Lin. 2018.
\newblock Unsupervised feature learning via non-parametric instance
  discrimination.
\newblock In \emph{Proceedings of the IEEE conference on computer vision and
  pattern recognition}, pages 3733--3742.

\bibitem[{Yu et~al.(2021)Yu, He, Zhang, Du, Pasupat, and
  Li}]{DBLP:conf/naacl/YuHZDPL21}
Dian Yu, Luheng He, Yuan Zhang, Xinya Du, Panupong Pasupat, and Qi~Li. 2021.
\newblock \href {https://doi.org/10.18653/v1/2021.naacl-main.59} {Few-shot
  intent classification and slot filling with retrieved examples}.
\newblock In \emph{Proceedings of the 2021 Conference of the North American
  Chapter of the Association for Computational Linguistics: Human Language
  Technologies, {NAACL-HLT} 2021, Online, June 6-11, 2021}, pages 734--749.
  Association for Computational Linguistics.

\bibitem[{Zhang et~al.(2021)Zhang, Nan, Wei, Li, Zhu, McKeown, Nallapati,
  Arnold, and Xiang}]{DBLP:conf/naacl/ZhangNWLZMNAX21}
Dejiao Zhang, Feng Nan, Xiaokai Wei, Shang{-}Wen Li, Henghui Zhu, Kathleen~R.
  McKeown, Ramesh Nallapati, Andrew~O. Arnold, and Bing Xiang. 2021.
\newblock \href {https://doi.org/10.18653/v1/2021.naacl-main.427} {Supporting
  clustering with contrastive learning}.
\newblock In \emph{Proceedings of the 2021 Conference of the North American
  Chapter of the Association for Computational Linguistics: Human Language
  Technologies, {NAACL-HLT} 2021, Online, June 6-11, 2021}, pages 5419--5430.
  Association for Computational Linguistics.

\bibitem[{Zhang et~al.(2017)Zhang, Sun, Eriksson, and Balzano}]{zhang2017deep}
Dejiao Zhang, Yifan Sun, Brian Eriksson, and Laura Balzano. 2017.
\newblock Deep unsupervised clustering using mixture of autoencoders.
\newblock \emph{arXiv preprint arXiv:1712.07788}.

\bibitem[{Zhang et~al.(2022)Zhang, Zhang, Zhan, Wu, and
  Lam}]{DBLP:conf/acl/ZhangZZ0L22}
Yuwei Zhang, Haode Zhang, Li{-}Ming Zhan, Xiao{-}Ming Wu, and Albert Y.~S. Lam.
  2022.
\newblock \href {https://doi.org/10.18653/v1/2022.acl-long.21} {New intent
  discovery with pre-training and contrastive learning}.
\newblock In \emph{Proceedings of the 60th Annual Meeting of the Association
  for Computational Linguistics (Volume 1: Long Papers), {ACL} 2022, Dublin,
  Ireland, May 22-27, 2022}, pages 256--269. Association for Computational
  Linguistics.

\bibitem[{Zhou et~al.(2022)Zhou, Zhang, Xiao, Dingwall, Ma, Arnold, and
  Xiang}]{DBLP:conf/naacl/0001Z0DMAX22}
Zhihan Zhou, Dejiao Zhang, Wei Xiao, Nicholas Dingwall, Xiaofei Ma, Andrew~O.
  Arnold, and Bing Xiang. 2022.
\newblock \href {https://doi.org/10.18653/v1/2022.naacl-main.55} {Learning
  dialogue representations from consecutive utterances}.
\newblock In \emph{Proceedings of the 2022 Conference of the North American
  Chapter of the Association for Computational Linguistics: Human Language
  Technologies, {NAACL} 2022, Seattle, WA, United States, July 10-15, 2022},
  pages 754--768. Association for Computational Linguistics.

\end{thebibliography}
\bibliographystyle{acl_natbib}

\appendix

\end{document}